\documentclass{llncs}
\usepackage{graphicx}
\usepackage[cp1250]{inputenc}
\usepackage{bm}
\usepackage{cite}
\usepackage{multirow}
\usepackage{mdwlist}
\usepackage{url}
\usepackage{arydshln}
\usepackage{algorithm}
\usepackage{algpseudocode}

\begin{document}

\title{Real-Time Hand Shape Classification}

\author{Jakub Nalepa\inst{1,2} \and Michal Kawulok\inst{1}}

\authorrunning{M. Kawulok and J. Nalepa}

\institute{Institute of Informatics, Silesian University of
Technology \\Akademicka 16, 44-100 Gliwice, Poland\\
\email{\{jakub.nalepa,michal.kawulok\}@polsl.pl}
\and Future Processing, Gliwice, Poland\\
\email{jnalepa@future-processing.com}}

\maketitle

\begin{abstract}
The problem of hand shape classification is challenging since a hand is characterized by a large number of degrees of freedom. Numerous shape descriptors have been proposed and applied over the years to estimate and classify hand poses in reasonable time. In this paper we discuss our parallel framework for real-time hand shape classification applicable in real-time applications. We show how the number of gallery images influences the classification accuracy and execution time of the parallel algorithm. We present the speedup and efficiency analyses that prove the efficacy of the parallel implementation. Noteworthy, different methods can be used at each step of our parallel framework. Here, we combine the shape contexts with the appearance-based techniques to enhance the robustness of the algorithm and to increase the classification score. An extensive experimental study proves the superiority of the proposed approach over existing state-of-the-art methods.
\end{abstract}

\section{Introduction}

Hand gestures constitute an important source of non-verbal communication, either complementary to the speech, or the primary one for people with disabilities. The problem of hand gesture recognition has been given a considerable research attention due to a wide range of its practical applications, including human-computer interfaces~\cite{Haq2011,CzuprynaELMAR2012}, virtual reality~\cite{Shen2011}, telemedicine~\cite{Tiwari2006}, videoconferencing~\cite{MacLean2001}, and more~\cite{GrzejszczakCORES2013,NalepaICMMI2014}. The proposed approaches can be divided into hardware- and vision- based methods. The former utilize sensors, markers and other equipment to deliver an accurate gesture recognition, but they lack naturalness and are of a high cost. Vision-based methods are contact-free, but require designing advanced image analysis algorithms for robust classification. Thus, an additional effort is needed for applying these techniques in real-time applications.

Numerous algorithms for hand shape classification have emerged over the years. In the contour-based approaches, the shape boundary of a detected hand is considered to represent its geometric features. These methods include, among others, a very time-consuming approach based on the shape contexts analysis and estimating similarity between shapes~\cite{Belongie2002}, recently optimized by reducing the search space by using the mean distances and standard deviations of shape contexts~\cite{Lin2011}, and Hausdorff distance-based methods~\cite{Huttenlocher1993}. The main drawback of these contour-based approaches lies in their limited use in case of missing contour information.

In the appearance-based methods, not only is the contour utilized for shape features extraction, but also the shape's internal region is analyzed. For example, an input color or greyscale image is processed in the orientation histograms approach~\cite{Freeman213} or an entire hand mask can be fed as an input to various template matching and moment-based methods~\cite{Thippur2013}. An interesting and thorough survey on vision-based hand pose estimation methods was published by Erol et al.~\cite{Erol2007}.

In this paper we discuss a fast parallel algorithm for hand shape classification. We show how the parallelization affects the classification time, and makes it possible to apply for searching large hand gesture sets in reasonable time. We present the speedup and efficiency of the parallel algorithm for various numbers of threads. Moreover, we show that combining the shape contexts with the appearance-based methods results in increasing the final classification score.

The paper is organized as follows. The hand shape classification algorithm is described in detail in Section~\ref{sec:parallel_algorithm}. The experimental study is reported in Section~\ref{sec:experimental_study} along with the description of the database of hand gestures. The paper is concluded and the directions of our future works are highlighted in Section~\ref{sec:conclusions}.

\section{Parallel Hand Shape Classification Algorithm}
\label{sec:parallel_algorithm}

\begin{figure}[t!]
\centering

\renewcommand{\tabcolsep}{0cm}
\newcommand{\myfigwidth}{0.23}
\newcommand{\raiseshift}{0.1mm}

\begin{tabular}{cccc}

\includegraphics[width=\myfigwidth\columnwidth]{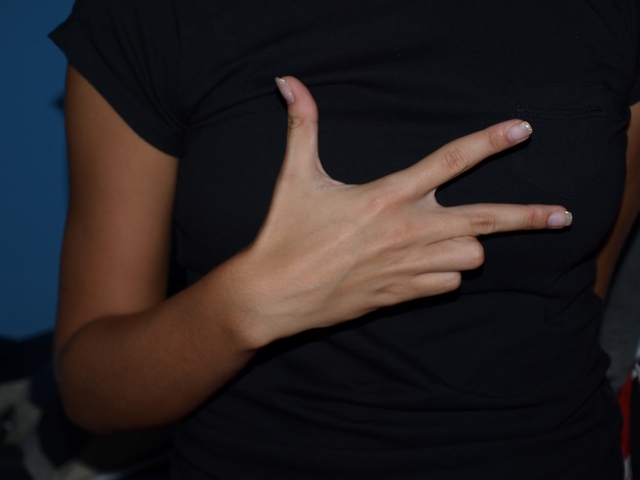} \hspace{0.01\columnwidth}&
\includegraphics[width=\myfigwidth\columnwidth]{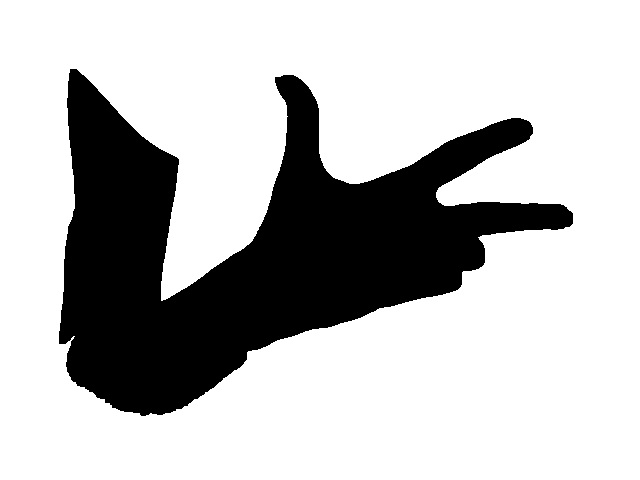} \hspace{0.01\columnwidth}&
\includegraphics[width=\myfigwidth\columnwidth]{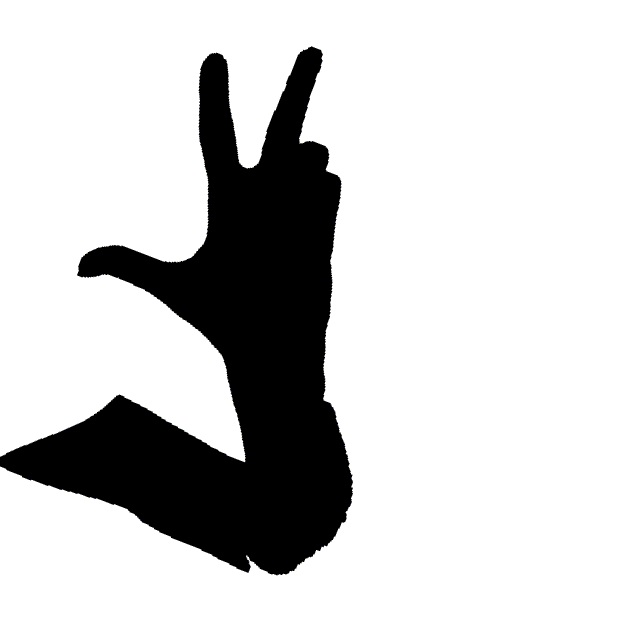} \hspace{0.01\columnwidth}&
\includegraphics[width=\myfigwidth\columnwidth]{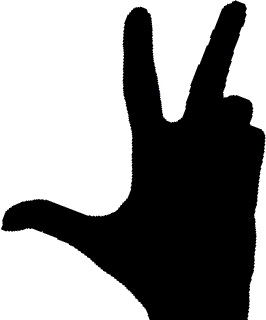}
\\[-1.0mm]

(A: input image) & (B: skin map) & (C: rotation) & (D: cropping)\\
\\[-1.8mm]

\end{tabular}

  \caption
  {
    Example of the hand skin map normalization.
  }
  \label{fig:ex_skin_map_normal}
\end{figure}

In this section we describe our parallel algorithm (PA) for hand shape classification~\cite{NalepaISM2013}. First, the input image $I_i$ is subject to skin segmentation, only if necessary (Alg.~\ref{alg:parallelClassification}, lines \ref{alg:verify_if_skin_analyzed}--\ref{alg:end_skin_detection}). This step is undertaken if the shape features are to be extracted from the skin map of $I_i$. There exist a number of robust skin detection and segmentation techniques~\cite{Jones2002, Kawulok2010MTaA, Kawulok2013ICIP, Kawulok2013FG, Kawulok2013}. A thorough survey on current state-of-the-art skin detection approaches has been published recently~\cite{Kawulok2013Springer}. Then, the image, either the skin mask or the original one, is normalized (line~\ref{alg:normalize_image}). The normalization procedure is presented in Fig.~\ref{fig:ex_skin_map_normal}. An input image (A) or the skin map (B) is rotated (C) based on the position of wrist points, so as the hand is oriented upwards. Pixels below the wrist line are discarded, the image is cropped and downscaled to the width $w_M$ (D).

Once the image is normalized, hand shape features are calculated in parallel (Alg.~\ref{alg:parallelClassification}, lines~\ref{alg:start_calculating_features}--\ref{alg:end_calculating_features}), and the input image $I_i$ is compared with the gallery images, also in parallel (lines~\ref{alg:start_classification}--\ref{alg:end_classification}). Finally, the classification result is returned (line~\ref{alg:return_final_classification}). Noteworthy, the classification procedure can be executed for a number of input images in parallel. Thus, larger databases of input hand images can be analyzed significantly faster than using a sequential approach.

\begin{algorithm}[t]
\algblock{ParFor}{EndParFor}
\algnewcommand\algorithmicparfor{\textbf{parfor}}
\algnewcommand\algorithmicpardo{\textbf{do}}
\algnewcommand\algorithmicendparfor{\textbf{end\ parfor}}
\algrenewtext{ParFor}[1]{\algorithmicparfor\ #1\ \algorithmicpardo}
\algrenewtext{EndParFor}{\algorithmicendparfor}
\centering
\begin{algorithmic}[1]
\ParFor{$I_i$ $\gets$ $I_1$ \textbf{to} $I_N$}\label{alg:start_calculating_features}
\State $SkinMapIsAnalyzed$ $\gets$ CheckIfSkinMapIsAnalyzed($I_i$);\label{alg:verify_if_skin_analyzed}
\If {$SkinMapIsAnalyzed$}\label{alg:is_skin_analyzed}
\State Detect skin and create skin map;\label{alg:detect_skin}
\EndIf\label{alg:end_skin_detection}
\State Normalize image;\Comment{See Fig.~\ref{fig:ex_skin_map_normal}}\label{alg:normalize_image}
\ParFor{$H_j$ $\gets$ $H_1$ \textbf{to} $H_M$}\label{alg:start_calculating_features}
\State Calculate j-$th$ hand shape feature;
\EndParFor \label{alg:end_calculating_features}
\ParFor{$G_j$ $\gets$ $G_1$ \textbf{to} $G_g$}\label{alg:start_classification}
\State Compare $I_i$ with j-$th$ gallery image;
\EndParFor \label{alg:end_classification}
\State \Return final hand shape classification;\label{alg:return_final_classification}
\EndParFor
\end{algorithmic}
  \caption
  {
    Parallel hand shape classification.
  }
  \label{alg:parallelClassification}
\end{algorithm}

In the shape features calculation and shape classification stages we utilized the following state-of-the-art methods: (1) shape contexts analysis (SC)~\cite{Belongie2002}, (2) template matching (TM), (3) Hausdorff distance analysis (HD)~\cite{Huttenlocher1993}, (4) comparison of the orientation histograms (HoG)~\cite{Freeman213}, (5) Hu moments analysis (HM)~\cite{Hu1962}, and two approaches combining the SC with the appearance-based methods: SC combined with the distance transform (SCDT) and SC enhanced by the orientation histograms analysis (SCH). In the TM algorithm we used the following summation for comparing the overlapped patches of images $I_1$ and $I_2$, of size $w_1 \times h_1$ and $w_2 \times h_2$, respectively:
\begin{equation} \label{eq:temp_match}
    R(i,j)=\sum_{a,b}(I_1(a,b) - I_2(i+a,j+b))^2
    \rm ,
\end{equation} where $0 \leq a \leq w_2-1$ and $0 \leq b \leq h_2 - 1$.

Let $\kappa$ and $\lambda$ be two contours compared in the SC method. For each point $p^{\kappa}_i$ and $p^{\lambda}_i$, $i\in\{1,\dots,m\}$, where $m$ is the number of contour points, belonging to $\kappa$ and $\lambda$ respectively, the coarse log-polar histogram $h_i$, i.e., the shape context, is calculated. It depicts the distribution of the remaining $(m-1)$ points for each $p_i$. Let $C_{ij}$ denote the cost of matching the points $p^{\kappa}_i$ and $p^{\lambda}_j$, given as a chi-square distance between the corresponding shape contexts. Then, the total matching cost of two contours $C$ is given as $C=\sum\nolimits_{i}C(p_i^\kappa,p_{\pi(i)}^\lambda)$, where $\pi$ is a permutation of the contour points. Clearly, the minimization of $C$ is an instance of the bipartite matching problem. It can be solved in $O(n^3)$ time, where $n$ is the number of sampled contour points, using the Hungarian method~\cite{Lin2011}. To speed up the SC, we sample and analyze a subset of $M_{SC}$, $M_{SC}\ll m$, contour points of a shape. Additionally, the distance transform (DT) of the hand mask from the contour is performed. Given the DT, its histogram $H_i$ is calculated for the image $I_i$. Then, the distance between the histograms $H_1$ and $H_2$ of two images $I_1$ and $I_2$ is found using the chi-square metric:
\begin{equation} \label{eq:hist_comp}
    d(H_1,H_2)=\sum_{B} \frac{(H_1(B)-H_2(B))^2}{H_1(B)+H_2(B)}
    \rm .
\end{equation} The final cost of shapes matching of the SCDT is given as: $C'=\alpha C+\beta d(H_1,H_2)$, where $\alpha$ and $\beta$ are the weights. Values of $C$ and $d(H_1,H_2)$ are normalized, therefore $0.0\leq C \leq 1.0$ and $0.0\leq d(H_1,H_2) \leq 1.0$. Similarly, the shape contexts are combined with the orientation histograms approach~\cite{Freeman213} using the same values of weights $\alpha$ and $\beta$.

\section{Experimental Results}
\label{sec:experimental_study}

The PA was implemented in C++ language using the OpenMP interface. The experiments were conducted on a computer equipped with an Intel Xeon 3.2 GHz (16 GB RAM with 6 physical cores and 12 threads) processor having the following cache hierarchy: 6 $\times$ 32 kB of L1 instruction and data cache, 6 $\times$ 256 kB L2 cache and 12 MB of L3 cache. The settings used in both stages of the PA were tuned experimentally to the following values: $\alpha=0.17$, $\beta=1.0$, $w_M=100$, $M_{SC}=20$.

\subsection{Database of Hand Gestures}
\begin{figure}[h!]
    \centering
    \begin{tabular}{cccccc}
		  & (I) & (II) & (III) & (IV) & (V) \\
(1) &
        \includegraphics[height=55px, width=0.19\columnwidth]{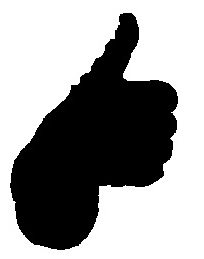} &
        \includegraphics[height=55px, width=0.19\columnwidth]{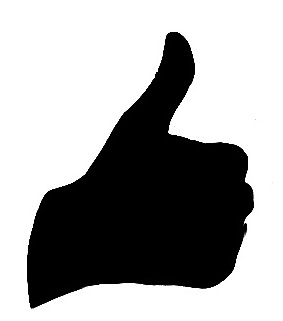} &
        \includegraphics[height=55px, width=0.19\columnwidth]{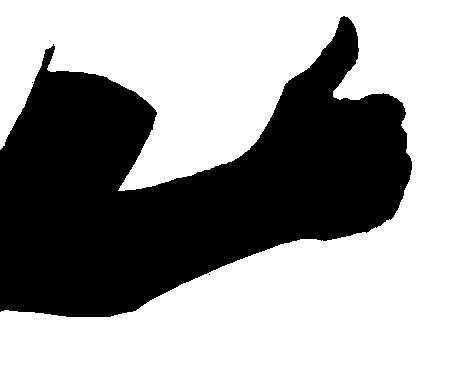} &
        \includegraphics[height=55px, width=0.19\columnwidth]{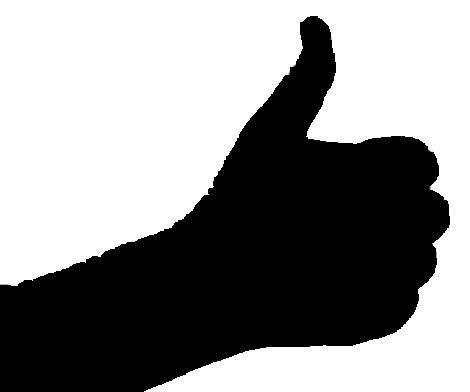} &
        \includegraphics[height=55px, width=0.19\columnwidth]{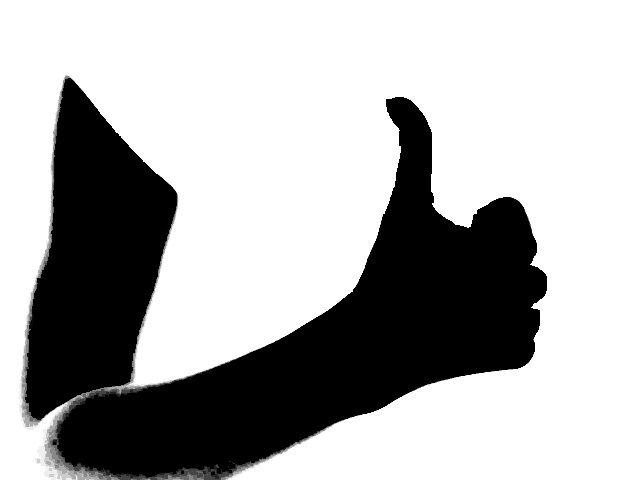} \\
(2) &
        \includegraphics[height=55px, width=0.19\columnwidth]{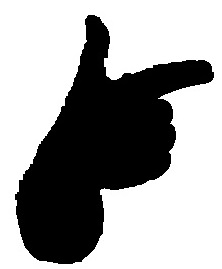} &
        \includegraphics[height=55px, width=0.19\columnwidth]{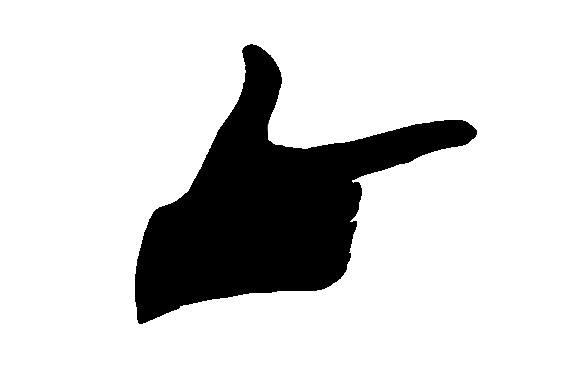} &
        \includegraphics[height=55px, width=0.19\columnwidth]{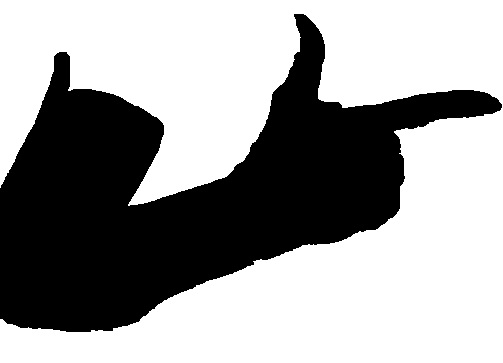} &
        \includegraphics[height=55px, width=0.19\columnwidth]{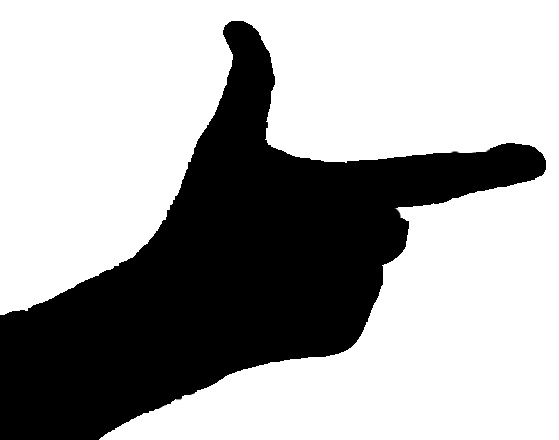} &
        \includegraphics[height=55px, width=0.19\columnwidth]{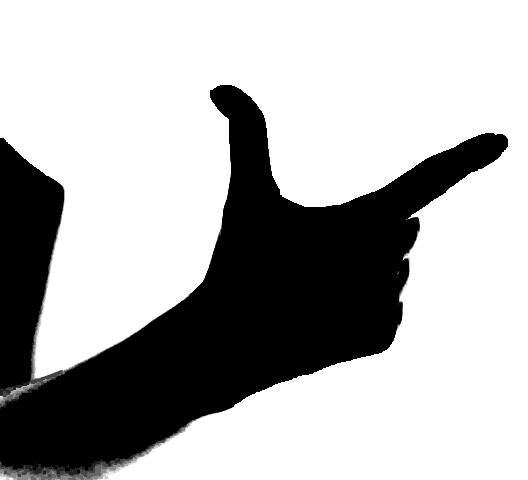} \\
(3) &
        \includegraphics[height=55px, width=0.19\columnwidth]{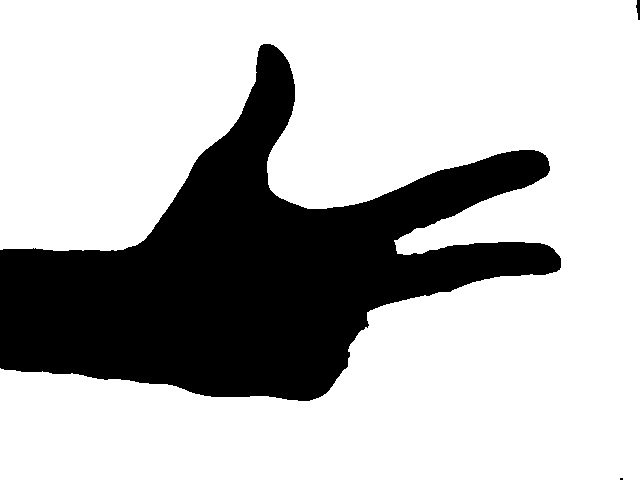} &
        \includegraphics[height=55px, width=0.19\columnwidth]{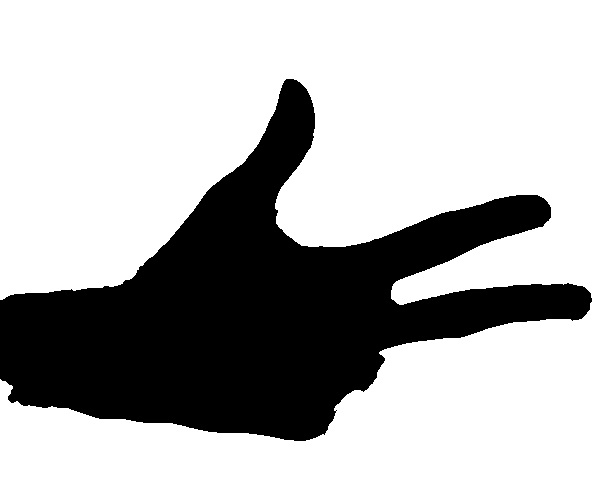} &
        \includegraphics[height=55px, width=0.19\columnwidth]{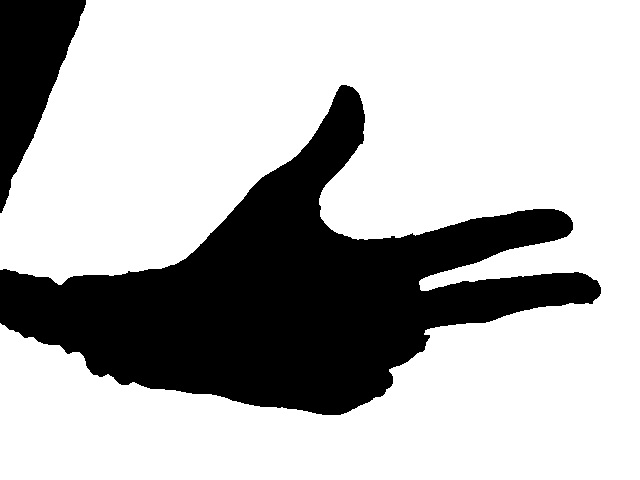} &
        \includegraphics[height=55px, width=0.19\columnwidth]{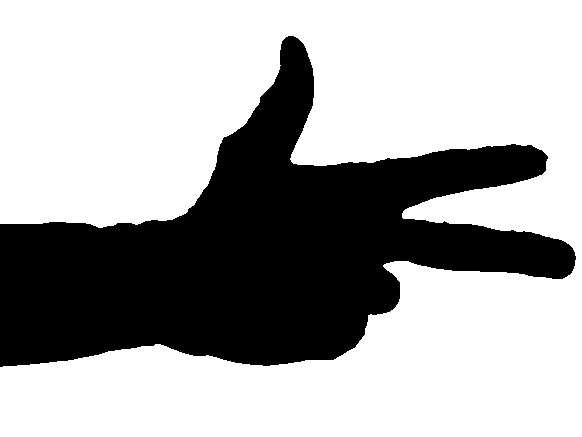} &
        \includegraphics[height=55px, width=0.19\columnwidth]{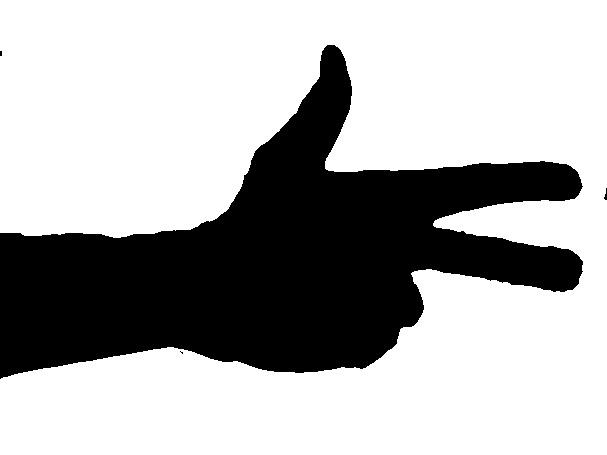} \\
(4) &
        \includegraphics[height=55px, width=0.19\columnwidth]{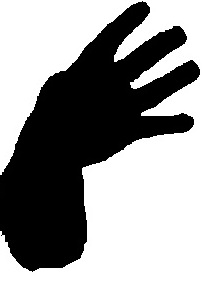} &
        \includegraphics[height=55px, width=0.19\columnwidth]{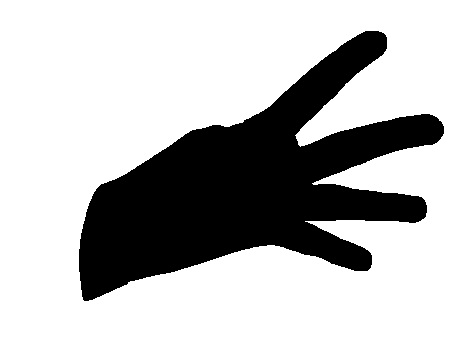} &
        \includegraphics[height=55px, width=0.19\columnwidth]{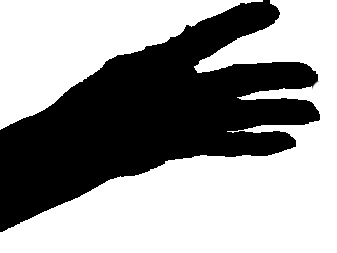} &
        \includegraphics[height=55px, width=0.19\columnwidth]{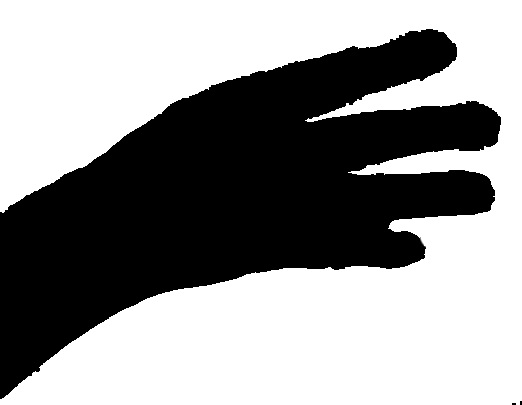} &
        \includegraphics[height=55px, width=0.19\columnwidth]{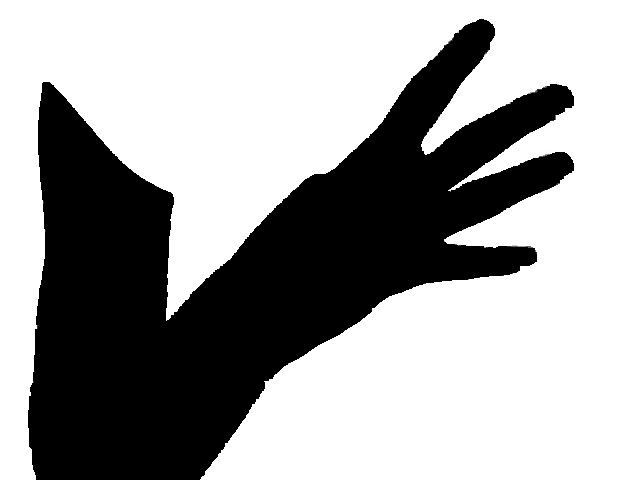} \\
(H) &
        \includegraphics[height=55px, width=0.19\columnwidth]{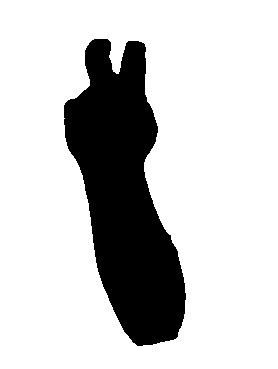} &
        \includegraphics[height=55px, width=0.19\columnwidth]{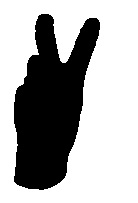} &
        \includegraphics[height=55px, width=0.19\columnwidth]{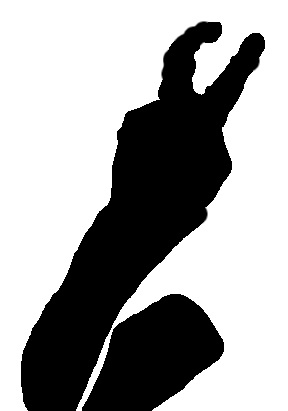} &
        \includegraphics[height=55px, width=0.19\columnwidth]{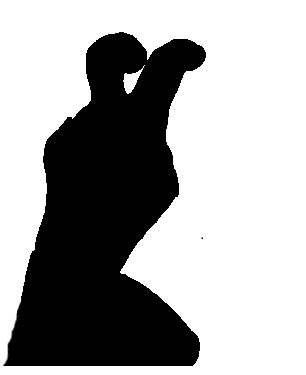} &
        \includegraphics[height=55px, width=0.13\columnwidth]{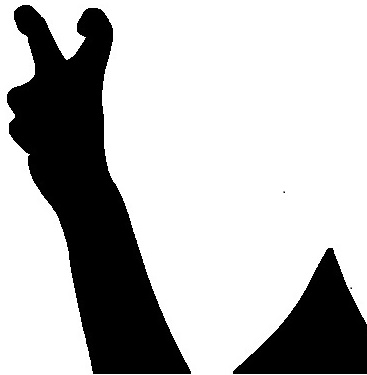} \\
(K) &
        \includegraphics[height=55px, width=0.19\columnwidth]{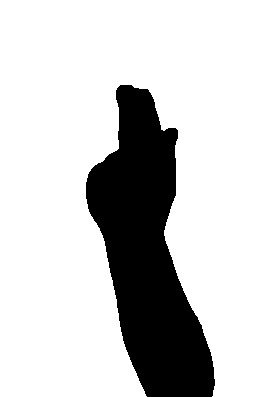} &
        \includegraphics[height=55px, width=0.19\columnwidth]{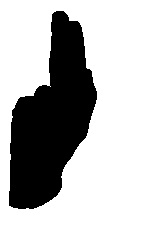} &
        \includegraphics[height=55px, width=0.19\columnwidth]{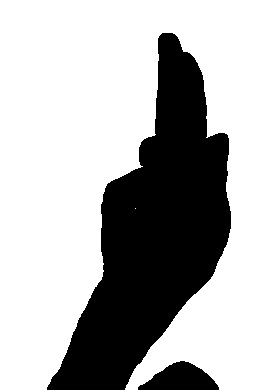} &
        \includegraphics[height=55px, width=0.19\columnwidth]{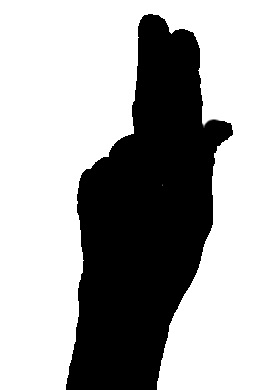} &
        \includegraphics[height=55px, width=0.15\columnwidth]{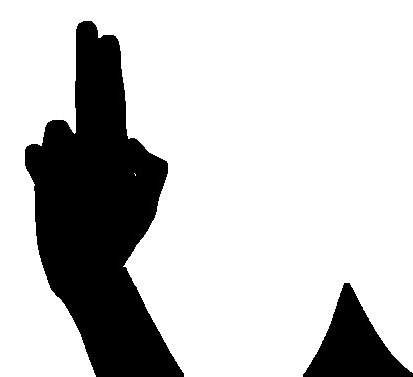} \\
(N) &
        \includegraphics[height=55px, width=0.19\columnwidth]{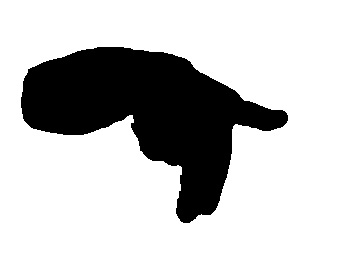} &
        \includegraphics[height=55px, width=0.19\columnwidth]{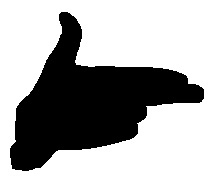} &
        \includegraphics[height=55px, width=0.19\columnwidth]{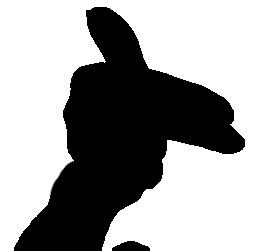} &
        \includegraphics[height=55px, width=0.19\columnwidth]{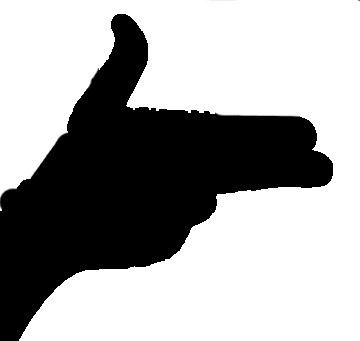} &
        \includegraphics[height=55px, width=0.13\columnwidth]{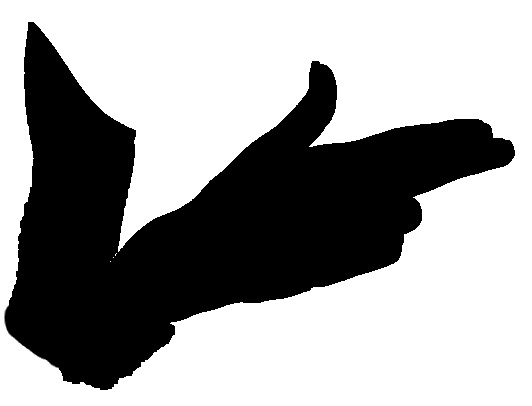} \\
(S) &
        \includegraphics[height=55px, width=0.19\columnwidth]{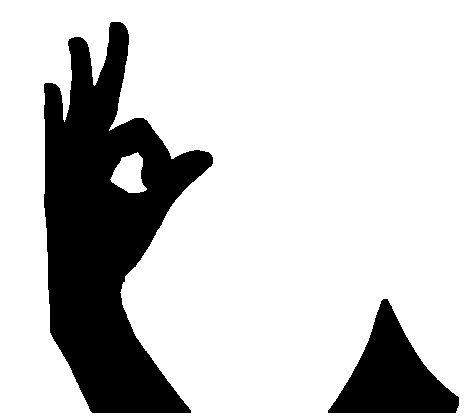} &
        \includegraphics[height=55px, width=0.19\columnwidth]{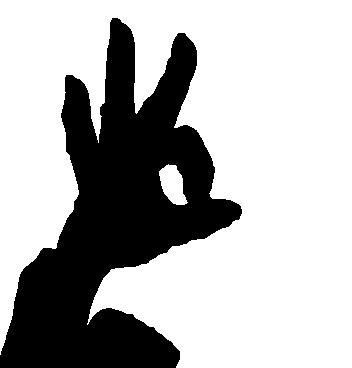} &
        \includegraphics[height=55px, width=0.19\columnwidth]{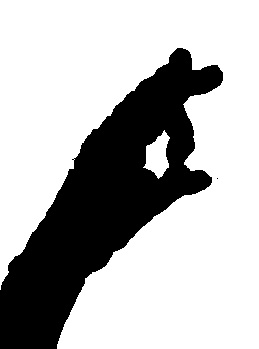} &
        \includegraphics[height=55px, width=0.19\columnwidth]{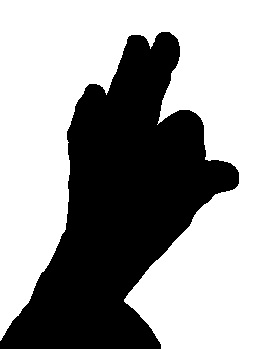} &
        \includegraphics[height=55px, width=0.13\columnwidth]{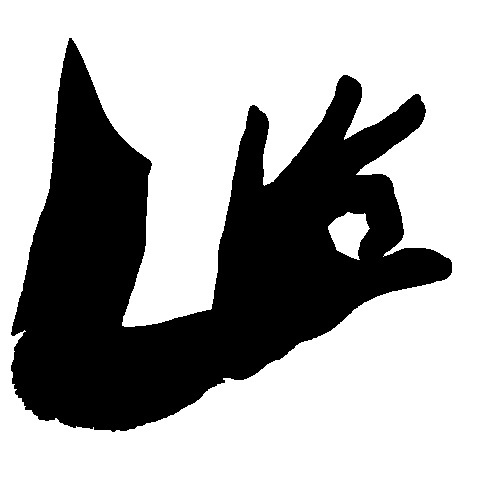}
    \end{tabular}
    \caption{Examples of ground-truth binary masks of various gestures (1, 2, 3, 4, H, K, N, S) presented by five individuals (I--V).}
    \label{jnalepa:fig:hand_examples}
\end{figure}

The experimental study was carried out using our database of 499 color hand images of 15 gestures presented by 12 individuals\footnote{For more details see \url{http://sun.aei.polsl.pl/~jnalepa/BDAS2014}}. Each gesture was presented $n$ times, $27\leq n \leq 39$. The images are associated with ground-truth binary masks indicating skin regions along with the ground-truth hand feature points. In this study, we omitted the skin segmentation and wrist localization stages, and used the ground-truth data for fair assessment of investigated techniques applied at other algorithm steps. Examples of ground-truth binary masks are presented in Fig.~\ref{jnalepa:fig:hand_examples}. Here, each gesture (1, 2, 3, 4, H, K, N, S) was presented by five individuals (I--V). It is easy to note that the difference between masks representing the same gesture (i.e.~inner-class difference) may be significant, e.g.~due to the hand rotation -- see e.g.,~Fig.~\ref{jnalepa:fig:hand_examples}(N).

\subsection{Classification Accuracy Analysis}

The data set was split into a gallery ($G$) and a probe set ($P$)~\cite{Phillips1998}. The gallery contains exactly $g$, $g\geq 1$, sample images per each gesture in the data set. Then, the similarities of the images from $P$ to those in $G$ were found using the techniques outlined in Section~\ref{sec:parallel_algorithm}. Classification effectiveness is assessed using its \emph{rank} ($R$), $1\leq R \leq\left\vert{G}\right\vert$. The rank is the position of the correct label on a list of gallery images sorted in the descending order by the similarity. If an image is classified correctly, then its rank is 1. The classification effectiveness for a given set is a percentage of correctly classified images. The analysis of the classification efficacy is performed based on cumulative response curves (CRCs).

\begin{figure}[h!]
\centering
\includegraphics[width=1\columnwidth]{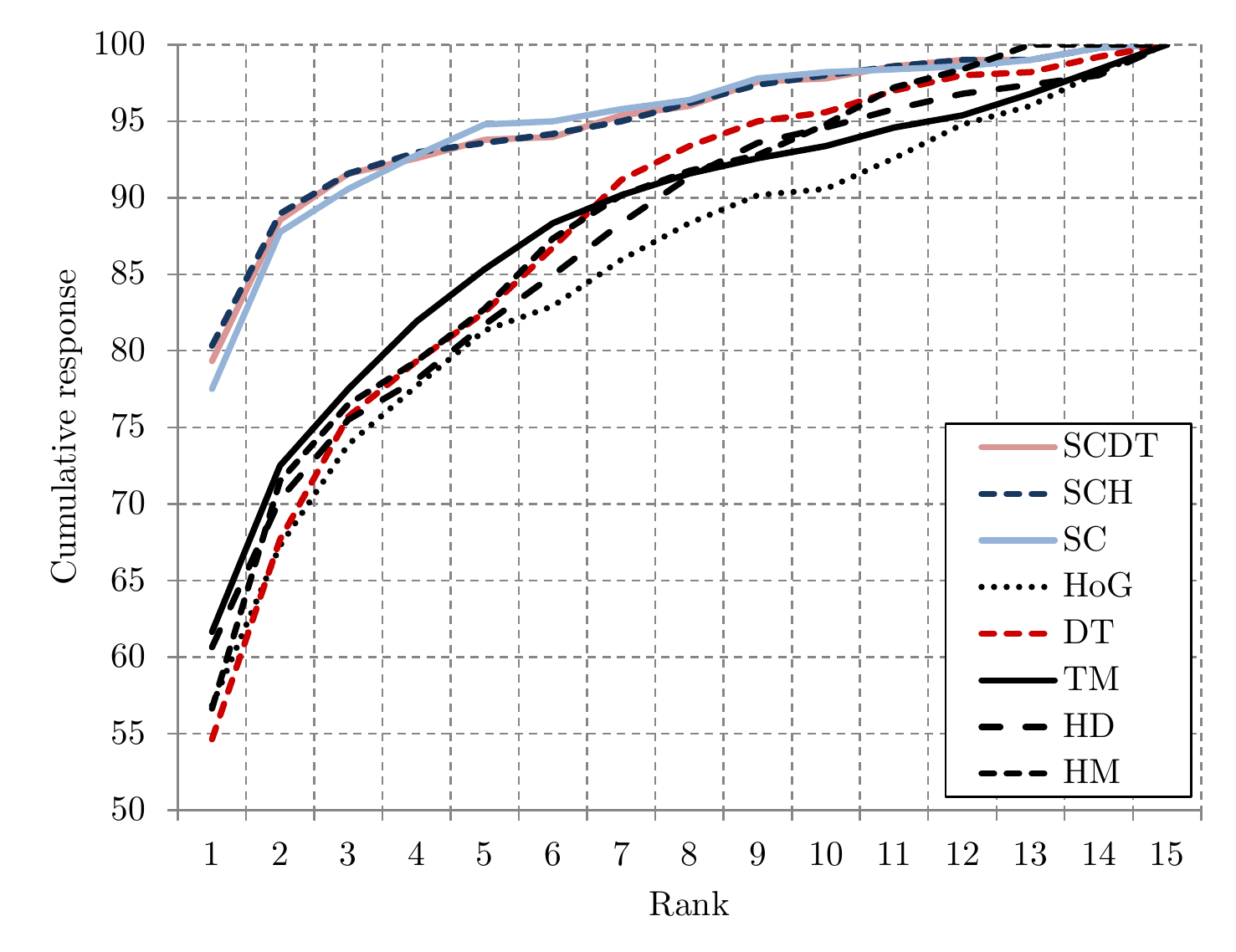}
  \caption
  {
    Best CRCs for a single image per gesture in $G$.
  }
  \label{fig:crc_gallery1_best}
\end{figure}

The best CRCs for a single image per gesture in $G$ are given in Fig.~\ref{fig:crc_gallery1_best}. We performed 27 classification experiments with no overlaps between the galleries. Then, the results were averaged and the best set of gallery images was determined. It is easy to see that the shape context methods, SCDT, SCH and SC, outperformed other techniques by at least ca.~15\%, considering the correct classification (see rank 1). Additionally, the algorithms enhanced by the appearance-based approaches, SCDT and SCH outperformed standard SC by 2\% and 3\%.

Tab.~\ref{tab:avg_crc_135image} shows the average CR values for 4 initial ranks along with their standard deviations $\sigma$. In the average case for a single image per gesture in $G$ (Tab.~\ref{tab:avg_crc_135image}(A)), it is the SCDT method which turned out to be the best among the investigated techniques, resulting in the highest CR values for each rank. Clearly, the choice of the image to $G$ has a strong impact on the later classification score, and selecting more distinctive images significantly affects the final results (see $\sigma$ in Tab~\ref{tab:avg_crc_135image}(A)). Although the standard deviation of the rank 1 is the smallest for the shape context based algorithms (SCDT, SCH, SC), it is still noticeable and proves the methods to be quite sensitive to the choice of the gallery images.

\begin{table}[t!]
  \caption{
    Average CR along with the standard deviation $\sigma$ (best CR shown in boldface) for various number of gallery images $g$: (A) $g=1$, (B) $g=3$, (C) $g=5$.
  }\label{tab:avg_crc_135image}

\renewcommand{\tabcolsep}{1mm}
\centering

\begin{tabular}{c c c: c c c c c c c c c }
\hline
&\multirow{2}{14mm}{Method~$\downarrow$} & & \multicolumn{1}{c}{Rank 1}  & & \multicolumn{1}{c}{Rank 2} & & \multicolumn{1}{c}{Rank 3} & & \multicolumn{1}{c}{Rank 4} \\
& & & CR $\pm$ $\sigma$ & & CR $\pm$ $\sigma$ & & CR $\pm$ $\sigma$& &CR $\pm$  $\sigma$\\
\hline

\multirow{8}{1mm}{(A)} & SCDT & & $\bf{69.30 \pm  5.84}$ & & $\bf{78.04 \pm  2.20}$ & & $\bf{82.92 \pm 1.80}$ & & $\bf{85.95 \pm 1.28}$ \\
&SCH  & & $69.02$ $\pm$  $6.33$ & & $77.82$ $\pm$  $2.19$ & & $82.38$ $\pm$  $1.82$ & & $85.40$ $\pm$  $1.17$ \\
&SC   & & $69.04$ $\pm$  $5.75$ & & $77.97$ $\pm$  $2.27$ & & $82.63$ $\pm$  $1.88$ & & $85.82$ $\pm$  $1.05$ \\
&HoG  & & $44.14$ $\pm$  $7.64$ & & $57.56$ $\pm$  $3.05$ & & $65.25$ $\pm$  $1.49$ & & $70.37$ $\pm$  $1.52$ \\
&DT  & & $41.95$ $\pm$  $6.71$ & & $56.87$ $\pm$  $2.76$ & & $66.34$ $\pm$  $1.86$ & & $72.95$ $\pm$  $1.47$ \\
&TM   & & $51.38$ $\pm$  $7.16$ & & $63.69$ $\pm$  $2.40$ & & $69.48$ $\pm$  $1.82$ & & $74.05$ $\pm$  $1.32$ \\
&HD   & & $49.10$ $\pm$  $7.44$ & & $59.81$ $\pm$  $2.01$ & & $66.39$ $\pm$  $1.62$ & &  $71.21$ $\pm$  $1.73$ \\
&HM   & & $46.69$ $\pm$  $7.07$ & & $60.86$ $\pm$  $3.48$ & & $68.46$ $\pm$  $2.52$ & & $74.57$ $\pm$  $1.62$ \\

\hdashline

\multirow{3}{1mm}{(B)} & SCDT & & $76.66$ $\pm$  $2.84$ & & $\bf{82.17 \pm 1.43}$ & & $85.71$ $\pm$  $0.90$ & & $\bf{88.47 \pm 1.19}$ \\
& SCH  & & $\bf{77.10 \pm 1.56}$ & & $82.11$ $\pm$  $0.87$ & & $\bf{86.23 \pm 0.61}$ & & $88.07$ $\pm$  $0.70$ \\
& SC   & & $75.74$ $\pm$  $2.45$ & & $81.00$ $\pm$  $1.47$ & & $84.23$ $\pm$  $0.88$ & & $86.75$ $\pm$  $0.73$ \\

\hdashline

\multirow{3}{1mm}{(C)} & SCDT & & $\bf{81.18 \pm 1.25}$ & & $85.97$ $\pm$  $0.87$ & & $\bf{88.35 \pm 0.74}$ & & $\bf{90.33 \pm 0.64}$ \\
& SCH  & & $80.59$ $\pm$  $2.30$ & & $\bf{86.06 \pm 1.38}$ & & $88.28$ $\pm$  $0.84$ & & $89.80$ $\pm$  $0.65$ \\
& SC   & & $80.17$ $\pm$  $1.69$ & & $85.18$ $\pm$  $1.57$ & & $87.63$ $\pm$  $0.86$ & & $89.21$ $\pm$  $0.55$ \\

\hline
\end{tabular}
\end{table}

\begin{figure}[h!]
\centering
\includegraphics[width=1\columnwidth]{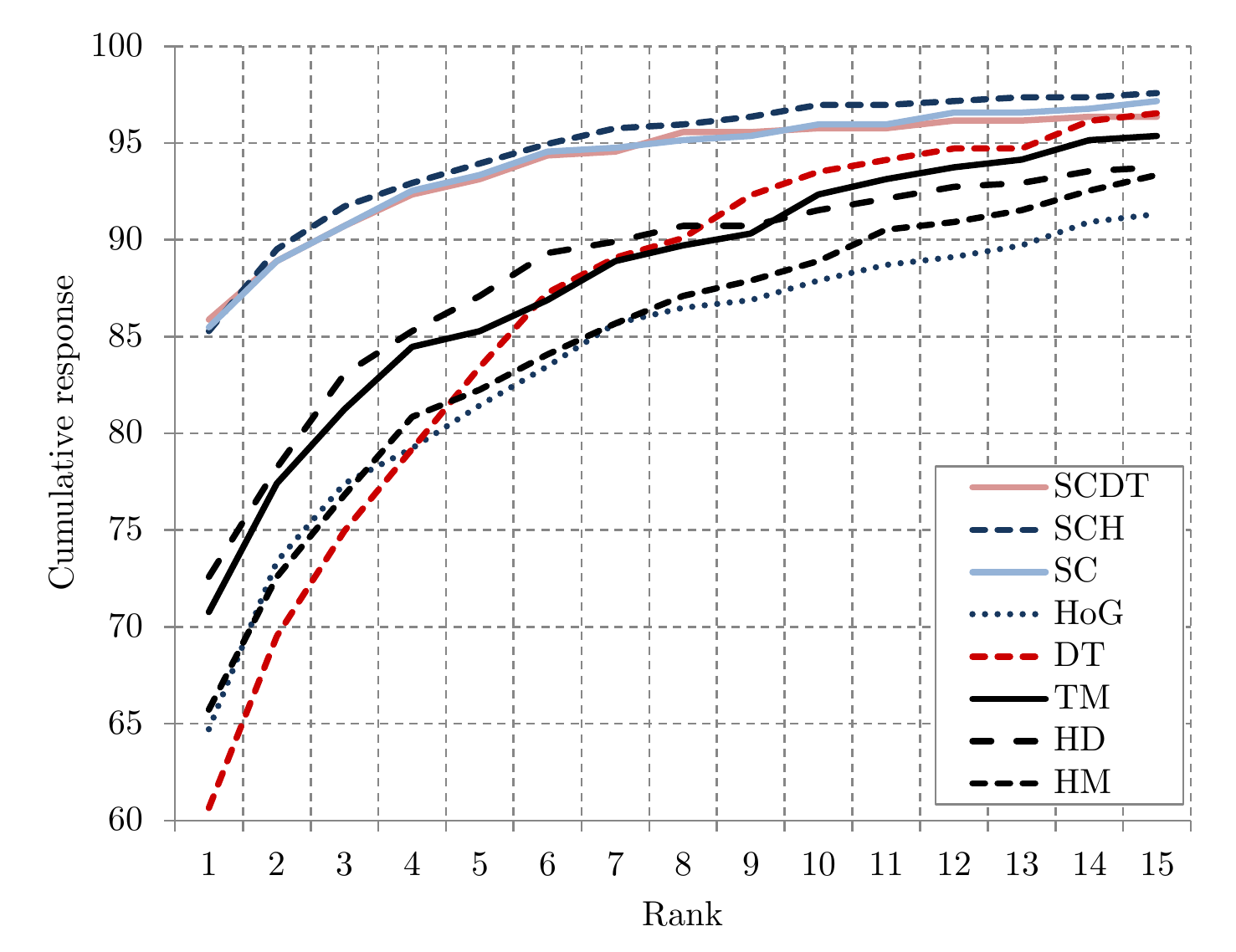}
  \caption
  {
    Best CRCs for three images per gesture in $G$.
  }
  \label{fig:crc_gallery3}
\end{figure}

Fig.~\ref{fig:crc_gallery3} presents the CRCs for three ($g=3$) most discriminative images, i.e.~these that gave the best score for $g=1$ for each method, per gesture in $G$. Providing multiple gallery entries improved the correct results in the initial ranks by at least 6\% (SCDT, SCH and DT), up to 12\% for the HD (see Fig.~\ref{fig:crc_gallery1_best} and Fig.~\ref{fig:crc_gallery3}). On the one hand, the appearance-based methods (HoG and DT) performed poorly for both $g=1$ and $g=3$. On the other hand, combining them with the contour-based shape contexts technique resulted in the best responses. Therefore, these methods are complementary. Noteworthy, combining the SC with other contour-based methods did not improve the classification score. Tab.~\ref{tab:avg_crc_135image}(B) and Tab.~\ref{tab:avg_crc_135image}(C) presents the average (out of 20 experiments) CR and its corresponding $\sigma$ for $g=3$ and $g=5$ for SC-based methods. The enhanced approaches outperformed the SC significantly. Moreover, adding new images to the gallery (i.e.~increasing $g$) made the algorithms more independent from the choice of gallery images (the $\sigma$ values dropped). 

\subsection{Speedup and Efficiency Analysis}

In order to assess the performance of the PA, we measured the analysis time $\tau(1)$ of the sequential algorithm and of the PA, $\tau(T)$, for various numbers of threads $T$, and calculated the speedup $\mathcal{S}=\tau(1)/\tau(T)$, along with the efficiency $E=\mathcal{S}/T$~\cite{Nalepa2012,Nalepa2013PPAM}. The analysis time $\tau$ consists of the feature extraction time $\tau_F$ and classification time $\tau_C$, thus $\tau=\tau_F+\tau_C$.

\begin{table}[t!]
  \caption{
    Average analysis time $\tau$, speedup $\mathcal{S}$ and efficiency $E$ of the PA for various numbers of threads $T$ and gallery images $g$: (A) $g=1$, (B) $g=3$, (C) $g=5$.
  }\label{tab:time_and_speedup}

\renewcommand{\tabcolsep}{1mm}
\centering

\begin{tabular}{c c c: c c c c c c c c c c c c c}
\hline
& \multirow{2}{14mm}{Method~$\downarrow$} & & $T=1$&  &\multicolumn{3}{c}{$T=2$} & & \multicolumn{3}{c}{$T=4$}& &\multicolumn{3}{c}{$T=8$} \\
& & & $\tau$ & & $\tau$ &$\mathcal{S}$ & $E$ & & $\tau$ & $\mathcal{S}$& $E$ & &$\tau$& $\mathcal{S}$ & $E$\\
\hline
\multirow{6}{1mm}{(A)} & SCDT & & $32$ & & $16$ & $1.96$ & $0.98$ & & $9$ & $3.45$ & $0.86$ & & $6$ & $5.22$ & $0.65$\\
& SCH  & & $39$ & & $19$ & $2.01$ & $1.01$ & & $11$ & $3.56$ & $0.89$ & & $7$ & $5.29$ & $0.66$\\
& SC   & & $32$ & & $16$ & $1.96$ & $0.98$ & & $9$ & $3.47$ & $0.87$ & & $6$ & $5.24$ & $0.66$\\
& HoG  & & $6$ & & $3$ & $1.76$ & $0.88$ & & $2$ & $3.08$ & $0.77$ & & $1$ & $4.68$ & $0.59$\\
& DT  & & $4$ & & $2$ & $1.70$ & $0.85$ & & $1$ & $3.02$ & $0.76$ & & $1$ & $4.61$ & $0.58$\\
& TM   & & $22$ & & $12$ & $1.80$ & $0.9$ & & $7$ & $3.28$ & $0.82$ & & $4$ & $5.11$ & $0.64$\\
& HD   & & $83$ & & $41$ & $2.05$ & $1.03$ & & $21$ & $3.89$ & $0.97$ & &  $14$ & $6.00$ & $0.75$\\
& HM & & $6$ & & $3$ & $1.88$ & $0.94$ & & $2$ & $3.39$ & $0.85$ & & $1$ & $4.84$ & $0.61$\\

\hdashline

\multirow{6}{1mm}{(B)} & SCDT & & $38$ & & $19$ & $1.99$ & $1.00$ & & $11$ & $3.43$ & $0.86$ & & $7$ & $5.46$ & $0.68$\\
& SCH   & & $45$ & & $22$ & $2.01$ & $1.01$ & & $12$ & $3.61$ & $0.90$& & $9$ & $5.17$ & $0.65$\\
& SC   & & $37$ & & $19$ & $1.97$ & $0.99$ & & $11$ & $3.41$ & $0.85$ & & $7$ & $5.36$ & $0.67$\\
& HoG  & & $6$ & & $3$ & $1.71$ & $0.86$ & & $2$ & $2.94$ & $0.74$ & & $1$ & $4.58$ & $0.57$\\
& DT  & & $4$ & & $2$ & $1.69$ & $0.85$ & & $1$ & $3.33$ & $0.83$ & & $1$ & $5.55$ & $0.69$\\
& TM   & & $54$ & & $31$ & $1.76$ & $0.88$ & & $17$ & $3.22$ & $0.81$ & & $11$ & $5.15$ & $0.64$\\
& HD   & & $218$ & & $115$ & $1.90$ & $0.95$ & & $64$ & $3.42$ & $0.86$ & &  $39$ & $5.59$ & $0.70$\\
& HM & & $6$ & & $3$ & $1.82$ & $0.91$ & & $2$ & $3.32$ & $0.83$ & & $1$ & $4.25$ & $0.53$\\

\hdashline

\multirow{3}{1mm}{(C)} & SCDT & & $52$ & & $25$ & $2.09$ & $1.05$ & & $13$ & $3.93$ & $0.98$ & & $9$ & $6.04$ & $0.76$\\
& SCH   & & $58$ & & $27$ & $2.15$ & $1.08$ & & $15$ & $3.86$ & $0.97$ & & $10$ & $5.84$ & $0.73$\\
& SC   & & $52$ & & $26$ & $1.96$ & $0.98$ & & $13$ & $3.84$ & $0.96$ & & $9$ & $5.89$ & $0.74$\\
\hline
\end{tabular}
\end{table}

We investigated the execution time of the sequential algorithm and the PA for both $g=1$ and $g=3$ for each technique. Also, we measured the analysis time for $g=5$ for the best approaches, i.e.~giving the best CR for a smaller number of images per gesture in $G$ (SCDT, SCH and SC). In the latter case, we run the PA 20 times, with 5 random images representing a given gesture, using each mentioned approach. The average analysis time $\tau$, along with the speedup $\mathcal{S}$ and efficiency $E$ of the PA for various number of parallel threads $T$, are shown in Tab.~\ref{tab:time_and_speedup}. The HD is the most time-consuming classification approach. Although we significantly reduced the number of contour points considered in the SC, SCDT and SCH techniques, their sequential analysis time is still relatively high. The HM, HoG and DT turned out to be very fast for both $g=1$ and $g=3$. Providing new images to $G$ increased the sequential analysis time of the TM and HD algorithms significantly.

The experiments performed for various number of threads $T$ showed that the sequential algorithm can be speeded up almost linearly in case of more computationally intensive approaches. It is worth to mention that we experienced the superlinear speedup~\cite{Chapman2007}, i.e.~$\mathcal{S}>T$ and $E>1.0$, while executing the PA with the HD, SCDT and SCH on two parallel threads ($T=2$). Our preliminary studies indicated the local core caches as the source of superlinearity, however this issue requires further investigation. Applying the PA allows for increasing the $G$ with a very fast analysis time and analyzing larger hand gesture databases. Thus, a more accurate classification can be performed in real-time (at more than $100$ frames per second rate) using the available processor resources, e.g.~the execution time of the SCDT ($g=5$, $T=8$) is more than 3.5 times lower than for a single thread and $g=1$ with a very significant increase in the classification score.

\section{Conclusions and Future Work}
\label{sec:conclusions}

In this paper we discussed our parallel algorithm for fast hand shape classification. Introducing the parallelism allowed for decreasing the execution time of the sequential algorithm significantly. Moreover, we showed that the classification score can be boosted without increasing the execution time if the available processor resources are utilized. We experienced the superlinear speedup, which indicates high efficacy of the parallelization. Furthermore, we presented how the selection of gallery images influences the classification score.

Our ongoing research includes increasing the classification accuracy of the proposed parallel algorithm. We consider using radial Chebyshev moments here as they occurred to be very effective for image retrieval purposes~\cite{Celebi2005}. Also, we plan to investigate the fusing schemes of contour-based and appearance-based techniques to enhance the final classification. Additionally, we aim at applying the proposed approach for searching the space of the parameters that control a 3D hand model~\cite{Saric2011}.

\bibliographystyle{splncs}
\bibliography{ref_all}

\begin{thebibliography}{10}

\bibitem{Haq2011}
Ul~Haq, E., Pirzada, S., Baig, M., Shin, H.:
\newblock New hand gesture recognition method for mouse operations.
\newblock In: Circuits and Systems (MWSCAS), 2011 IEEE 54th International
  Midwest Symposium on. (2011)  1--4

\bibitem{CzuprynaELMAR2012}
Czupryna, M., Kawulok, M.:
\newblock Real-time vision pointer interface.
\newblock In: ELMAR, 2012 Proceedings. (2012)  49--52

\bibitem{Shen2011}
Shen, Y., Ong, S.K., Nee, A.Y.C.:
\newblock Vision-based hand interaction in augmented reality environment.
\newblock Int. J. Hum. Comput. Interaction \textbf{27}(6) (2011)  523--544

\bibitem{Tiwari2006}
Wachs, J., Stern, H., Edan, Y., Gillam, M., Feied, C., Smith, M., Handler, J.:
\newblock A real-time hand gesture interface for medical visualization
  applications.
\newblock In: App. of Soft Comp. Volume~36.
\newblock Springer Berlin Heidelberg (2006)  153--162

\bibitem{MacLean2001}
MacLean, J., Pantofaru, C., Wood, L., Herpers, R., Derpanis, K., Topalovic, D.,
  Tsotsos, J.:
\newblock Fast hand gesture recognition for real-time teleconferencing
  applications.
\newblock In: Proc. IEEE ICCV Workshop on Recognition, Analysis, and Tracking
  of Faces and Gestures in Real-Time Systems. (2001)  133--140

\bibitem{GrzejszczakCORES2013}
Grzejszczak, T., Nalepa, J., Kawulok, M.:
\newblock Real-time wrist localization in hand silhouettes.
\newblock In Burduk, R., Jackowski, K., Kurzynski, M., Wozniak, M., Zolnierek,
  A., eds.: Proceedings of the 8th International Conference on Computer
  Recognition Systems CORES 2013. Volume 226 of Advances in Intelligent Systems
  and Computing.
\newblock Springer International Publishing (2013)  439--449

\bibitem{NalepaICMMI2014}
Nalepa, J., Grzejszczak, T., Kawulok, M.:
\newblock Wrist localization in color images for hand gesture recognition.
\newblock In Gruca, A., Czachorski, T., Kozielski, S., eds.: Man-Machine
  Interactions 3. Volume 242 of Advances in Intelligent Systems and Computing.
\newblock Springer International Publishing (2014)  79--86

\bibitem{Belongie2002}
Belongie, S., Malik, J., Puzicha, J.:
\newblock Shape matching and object recognition using shape contexts.
\newblock IEEE TPAMI \textbf{24}(4) (2002)  509--522

\bibitem{Lin2011}
Lin, C.C., Chang, C.T.:
\newblock A fast shape context matching using indexing.
\newblock In: Proc. IEEE ICGEC. (2011)  17--20

\bibitem{Huttenlocher1993}
Huttenlocher, D., Klanderman, G., Rucklidge, W.:
\newblock Comparing images using the hausdorff distance.
\newblock IEEE TPAMI \textbf{15}(9) (1993)  850--863

\bibitem{Freeman213}
Freeman, W.T., Roth, M.:
\newblock Orientation histograms for hand gesture recognition.
\newblock Technical report, MERL (1994)

\bibitem{Thippur2013}
Thippur, A., Ek, C.H., Kjellstrom, H.:
\newblock Inferring hand pose: A comparative study of visual shape features.
\newblock In: Proc. IEEE FG. (2013)  1--8

\bibitem{Erol2007}
Erol, A., Bebis, G., Nicolescu, M., Boyle, R.D., Twombly, X.:
\newblock Vision-based hand pose estimation: A review.
\newblock Comp. Vis. and Im. Underst. \textbf{108}(1–2) (2007)  52--73

\bibitem{NalepaISM2013}
Nalepa, J., Kawulok, M.:
\newblock Parallel hand shape classification.
\newblock In: Proc. IEEE ISM. (2013)  401--402

\bibitem{Jones2002}
Jones, M., Rehg, J.:
\newblock Statistical color models with application to skin detection.
\newblock International J. of Computer Vis. \textbf{46} (2002)  81--96

\bibitem{Kawulok2010MTaA}
Kawulok, M.:
\newblock Energy-based blob analysis for improving precision of skin
  segmentation.
\newblock Multimedia Tools and Applications \textbf{49}(3) (2010)  463--481

\bibitem{Kawulok2013ICIP}
Kawulok, M., Kawulok, J., Nalepa, J., Papiez, M.:
\newblock Skin detection using spatial analysis with adaptive seed.
\newblock In: Proc. IEEE ICIP. (2013)  3720–3724

\bibitem{Kawulok2013FG}
Kawulok, M.:
\newblock Fast propagation-based skin regions segmentation in color images.
\newblock In: Proc. IEEE FG. (2013)  1--7

\bibitem{Kawulok2013}
Kawulok, M., Kawulok, J., Nalepa, J.:
\newblock Spatial-based skin detection using discriminative skin-presence
  features.
\newblock Pattern Recognition Letters (0) (2013) ~-- in press.

\bibitem{Kawulok2013Springer}
Kawulok, M., Nalepa, J., Kawulok, J.:
\newblock Skin detection and segmentation in color images.
\newblock In Celebi, M.E., Smolka, B., eds.: Advances in Low-Level Color Image
  Processing. Volume~11 of Lecture Notes in Computational Vision and
  Biomechanics.
\newblock Springer Netherlands (2014)  329--366

\bibitem{Hu1962}
Hu, M.K.:
\newblock Visual pattern recognition by moment invariants.
\newblock Inf. Theory, IRE Trans. on \textbf{8}(2) (1962)  179--187

\bibitem{Phillips1998}
Phillips, P., Wechsler, H., Huang, J., Rauss, P.:
\newblock The {FERET} database and evaluation procedure for face recognition
  algorithms.
\newblock Im. and Vis. Comp. J. \textbf{16}(5) (1998)  295--306

\bibitem{Nalepa2012}
Nalepa, J., Czech, Z.J.:
\newblock A parallel heuristic algorithm to solve the vehicle routing problem
  with time windows.
\newblock Studia {I}nformatica \textbf{33}(1) (2012)  91--106

\bibitem{Nalepa2013PPAM}
Nalepa, J., Blocho, M., Czech, Z.J.:
\newblock Co-operation schemes for the parallel memetic algorithm.
\newblock In: Parallel Processing and Applied Mathematics. Lecture Notes in
  Computer Science.
\newblock Springer Berlin Heidelberg (2013) in press.

\bibitem{Chapman2007}
Chapman, B., Jost, G., Pas, R.v.d.:
\newblock {Using OpenMP: Portable Shared Memory Parallel Programming}.
\newblock The MIT Press (2007)

\bibitem{Celebi2005}
Celebi, M., Aslandogan, Y.:
\newblock A comparative study of three moment-based shape descriptors.
\newblock In: Proc. IEEE ITCC. Volume~1. (2005)  788--793

\bibitem{Saric2011}
\v{S}ari\'{c}, M.:
\newblock Libhand: A library for hand articulation (2011) Version 0.9.

\end{thebibliography}
\end{document}